\documentclass{article}

 \usepackage[preprint, nonatbib]{neurips_2026}

\usepackage[utf8]{inputenc} 
\usepackage[T1]{fontenc}    
\usepackage{hyperref}       
\usepackage{url}            
\usepackage{booktabs}       
\usepackage{amsfonts}       
\usepackage{nicefrac}       
\usepackage{microtype}      
\usepackage{xcolor}         
\usepackage{graphicx}
\usepackage{amsmath}
\usepackage{multirow}

\title{Selector-Guided Autonomous Curriculum for One-Shot Reinforcement Learning from Verifiable Rewards}

\author{%
  Rudray Dave \\
  Department of Artificial Intelligence\\
  Sardar Vallabhbhai National Institute of Technology \\
  Surat, India \\
  \texttt{U23AI123@coed.svnit.ac.in} \\
  \And
  Vedang Dubey\thanks{Authors have done equal contribution} \\
  Department of Artificial Intelligence\\
  Sardar Vallabhbhai National Institute of Technology\\
  Surat, India \\
  \texttt{U23AI111@coed.svnit.ac.in} \\
\AND
  Smit Deoghare\footnotemark[1] \\
  Department of Artificial Intelligence\\
  Sardar Vallabhbhai National Institute of Technology\\
  Surat, India \\
  \texttt{U23AI118@coed.svnit.ac.in} \\
\And
  Sudhakar ~Mishra\thanks{Corresponding Author: sudhakarm@aid.svnit.ac.in} \\
    Department of Artificial Intelligence\\
  Sardar Vallabhbhai National Institute of Technology\\
  Surat, India \\
  \texttt{sudhakarm@aid.svnit.ac.in} \\
}

\begin{document}

\maketitle

\begin{abstract}
  Recently, Reinforcement Learning from Verifiable Rewards (RLVR) has been established as a highly effective technique for augmenting the math reasoning skills of Large Language Models (LLMs) based on a single instance. Current state-of-the-art 1-shot RLVR models adopt heuristics for selecting instances, mostly based on historical variance in rewards, which we find to be inherently misleading as a measure of transferability value. In this paper, we propose a Selector-Guided Autonomous Curriculum (SGAC) approach, which employs a learnable selector model on a multi-dimensional feature space consisting of success probability, reward variance, output disagreement (entropy), and semantic difficulty level, instead of the static reward variance heuristic. In our empirical evaluation on pools of candidate problems, we observe that output disagreement, rather than reward variance, is the strongest predictor of reasoning gains in subsequent iterations. Leveraging this finding, we develop an autonomous curriculum algorithm for dynamically siphoning candidate problems from a large pool, ranking them by the learned selector, and running micro-bursts of 1-shot GRPO. Our framework is evaluated using the Hendrycks MATH benchmark, with the Qwen2.5-Math-1.5B model serving as the baseline. Our framework obtains an accuracy of 68.0\% on the hold-out dataset, which is better than the accuracy obtained from the baseline model, 64.0\%, as well as the 1-shot RLVR checkpoint proposed by Wang et al., which achieved an accuracy of 66.0\%. The results confirm that entropy-based intelligent data curation leads to strict reasoning improvement over static training methods, particularly in severely limited data conditions.
\end{abstract}

\section{Introduction}
\label{sec:introduction}

Large Language Models (LLMs) have proven their proficiency in a wide range of natural language processing tasks, including code generation, question answering, and logical reasoning \cite{b1, b2}. Yet, mathematical reasoning continues to be one of the most difficult fronts, entailing multi-step logical deductions, symbol manipulations, and problem solving beyond mere pattern recognition \cite{b1}. Thus, advancing mathematical reasoning in LLMs is an important direction of research.

Recently, a new approach has been developed for enhancing the reasoning ability of LLMs through Reinforcement Learning from Verifiable Rewards (RLVR) \cite{b3, b4}. Compared to reinforcement learning using costly and ambiguous human preferences as input (i.e., RLHF), the verifiability of mathematical problems gives rise to an objective reward function. Specifically, the outputs of an LLM can only be right or wrong, providing a clear binary reward. This makes RLVR more suitable than its alternative for training LLMs on mathematical reasoning.

One interesting discovery from the RLVR literature is that of Wang et al. \cite{b5}, which proved that it is possible to make significant improvements in the mathematical reasoning ability of an LLM by training it with RLVR with just one or two training examples. In their paper, titled "Reinforcement Learning for Reasoning in Large Language Models with One Training Example," Wang et al. proved that it is possible to enhance the reasoning abilities of the model by using the same problem repeatedly with the Group Relative Policy Optimization (GRPO) \cite{b6} algorithm. This 1-shot RLVR approach is surprising since it contradicts the commonly held notion that large and well-curated data sets are required for capability enhancements.

The surprising insight offered by Wang et al. is that the model learns reasoning skills that can be transferred to solving other mathematical tasks by training on just one such task in many GRPO repetitions. The authors’ assumption is that, through reinforcement learning, the model is not made able to solve the particular task, but its reasoning capabilities that had been learned during the pre-training stage and were not being activated during normal inference are brought to life. Such capabilities include self-examination, step-by-step validation, and systematic thinking. The RL training process is a "catalyst".

\subsection{Problem Statement}

Even with the potential of 1-shot RLVR, an outstanding issue still exists: \textbf{how to determine a "good" example in the context of RLVR?} In the study by Wang et al. \cite{b5}, they used a strategy involving selecting problems that had high variance in their rewards over many instances of the same problem generated by the model. The idea is that the problems that have high variance would represent regions of the model's capability space wherein the model is "uncertain," which means that resolving its uncertainties using reinforcement learning will yield maximum downstream transfer effects.

Unfortunately, this strategy is inherently flawed. Using empirical analyses, we show that high variance in reward scores often comes from extraneous factors like errors in parsing, tokenizing, and formatting that do not reflect any actual uncertainty in the concept. Reward variance, being a scalar quantity representing several different sources of variation, is thus an inadequate predictor of the learning potential of a particular problem.

The selection of appropriate training examples is particularly critical for the 1-shot RLVR setting for a number of reasons. Firstly, the reason being that a single example is used, the significance of the decision becomes highly pronounced, as there is no diversification of risk by using more examples. Secondly, this selection occurs only once, before even the training process starts and there is no possibility of changing it midway during the learning process. Lastly, due to the large number of possibilities in the selection (such as 12,000 examples in the MATH dataset), testing all of them is not feasible.

\subsection{Our Solution}

We introduce a data-driven methodology for training an example selector model, which uses an optimal set of features based on a rigorous learning methodology rather than variance heuristics. The proposed methodological solution has been developed as a three-step process, comprising:

\begin{enumerate}
    \item \textbf{Variance Fallacy}: We systematically demonstrate that reward variance is not a reliable predictor of downstream transfer accuracy. In our controlled experiments, the candidate problem with the maximum reward variance (0.250) led to the \textit{least} downstream accuracy (40\%), which clearly contradicts the existing heuristic. This result casts doubt on a core assumption in the data curation of RLVR studies.

    \item \textbf{Disagreement as Primary Catalyst}: We find that the output disagreement (the entropy of unique mathematical answers from \(K \)stochastic rollouts) serves as a much stronger predictor than variance. A candidate with 100\% disagreement, where all rollouts generated unique answers, achieved the best downstream accuracy (50\%). We present theoretical justifications for why output disagreement captures mathematical-level uncertainty and reward variance captures uncertainty and measurement error.

    \item \textbf{Learned Selector Model}: We learn a Linear Regression selector model from empirically measured four dimensions—success probability, reward variance, disagreement, and semantic difficulty level—as features for predicting downstream transfer accuracy. We discover that success probability has a negative coefficient ($w=-0.0574$) since easily solvable tasks do not pose a significant challenge, whereas difficulty level has a positive coefficient ($w=+0.1095$).
    
    \item \textbf{Fully Autonomous Curriculum Loop}: The learned selector is incorporated into an entirely autonomous training pipeline where: (a) candidate tasks are randomly sampled from a vast library of problems, (b) multiple signals are computed for each candidate using stochastic generation, (c) the best candidate is selected by the learned selector, and (d) a targeted GRPO micro-burst is conducted on the selected problem. Through 20 curriculum steps with accumulated LoRA weights, the autonomous pipeline obtains 68.0\% accuracy, outperforming the baseline model (64.0\%) and the original RLVR checkpoint (66.0\%).
\end{enumerate}

Not only the practical advancement, but also the theoretical breakthrough that our work implies should be emphasized: rather than seeing data selection as an already optimized heuristic task, we view it as a \textit{learnable optimization objective} which could potentially achieve systematic improvements. The ability to outperform heuristically-designed solutions even by very simple models such as linear models under proper training signals is shown by our framework.

The rest of this paper is structured as follows: We review relevant literature on RLVR, curriculum learning and data selection methods for LLMs in Section~\ref{sec:related}. Our methodology, including selector model training, autonomous curriculum loop construction and experiments design, is introduced in Section~\ref{sec:methodology}. Experimental results and comparison with other state-of-the-art approaches and ablation studies are provided in Section~\ref{sec:results}. Section~\ref{sec:discussion} discusses the implications, limitations and future works of our approach. Section~\ref{sec:conclusion} concludes the paper.

\section{Related Work}
\label{sec:related}

\subsection{Reinforcement Learning from Verifiable Rewards}

The idea of employing verifiable reward signals for reinforcement learning in language models has been gaining considerable attention lately. In particular, DeepSeek-R1 \cite{b7} proved that RL training alone without any further supervised fine-tuning could generate models that exhibit excellent reasoning abilities. More specifically, DeepSeek-R1 found that the emergence of behaviors like self-verifying and self-reflection in RL-trained models was a result of their capacity for reasoning. Most importantly, DeepSeek-R1 found that such reasoning behaviors emerge spontaneously without having to train the models on any reasoning traces; instead, reasoning emerges due to the nature of RL.

The Group Relative Policy Optimization (GRPO) \cite{b6} technique, presented as part of the DeepSeekMath model, obviates the necessity of training a separate critic model by estimating relative rewards based on multiple samples generated for the same prompt. It is more memory-efficient since no extra value network needs to be maintained and trained along with the language model during training. The GRPO advantage for the \(i$-th completion among \(G \)completions is calculated as:

\begin{equation}
    \hat{A}^{\text{GRPO}}_i = \frac{R_i - \mu_G}{\sigma_G + \epsilon}
\end{equation}

with \(\mu_G \)and \(\sigma_G \)being the mean and standard deviation of reward values in the group, respectively. Such a relative scaling of the advantage function prevents its bias due to the absolute level of the reward and guarantees balance between reinforcement and punishment.

One-Shot RLVR \cite{b5} by Wang et al. revealed the remarkable capability of GRPO when learning from a single experience, yielding improvements similar to those obtained with significantly larger amounts of data. The reason why multiple iterations of RL training on one problem do not result in overfitting, but in uncovering general reasoning capabilities, was the main discovery made by the authors.

\subsection{Curriculum Learning}

Curriculum learning \cite{b8} suggests ordering training data from easiest to hardest to enable better learning efficiency. The basic concept behind curriculum learning stems from education: similarly to how humans benefit from the process of gradually learning new information in a systematic way, artificial intelligence models could be trained with increasing complexity. For language models, curriculum-based techniques have been utilized for ordering of training data during pre-training \cite{b9}, fine-tuning of LLMs, and reward shaping in reinforcement learning. Self-paced learning \cite{b10} goes further in incorporating the principle of curriculum learning: the model decides which instances to train on according to its skill level.

Anti-curriculum methods \cite{b22}, where the models start training from the most challenging tasks, have been proved to perform well in some cases. Unlike traditional anti-curricula and curricula, our autonomous curriculum is neither fully \textit{easy-to-hard} or \textit{hard-to-easy}. Rather, we refer to our algorithm as \textit{informativeness-driven}: at each step, our algorithm chooses the problem that maximizes downstream utility predicted by the learned selector.

Our own curriculum generation loop shares conceptual similarity with both techniques, yet there is one major difference between them: Instead of having the next sequence of examples already determined, we \textit{choose} the best possible example at each stage from the pool generated by our dynamic sampling strategy, based on how useful each of them might be for the model down the road, as captured in our trained selector. Our choice is guided by an instantaneous measurement of uncertainty about the candidate examples.

\subsection{Data Selection and Active Learning}

The process of selecting data for machine learning has been studied extensively \cite{b11}. Active learning strategies generally rely on the selection of examples with maximum uncertainty according to some metric, such as entropy or query-by-committee differences \cite{b12}. Our proposed criterion for selecting examples can be considered similar to the query-by-committee methodology, since both rely on analyzing the outputs of the model ("committee members") for consensus. The novelty of our proposal is in applying the method to autoregressive models, where the "members" refer not to the individual models themselves but rather to multiple realizations of random rollouts sampled from the output of one model.

Several techniques have recently been applied in the context of data selection for the training of large language models, including methods based on influence functions \cite{b13}, gradient-based selection \cite{b14}, and perplexity-based filtering \cite{b15}. These approaches all pertain to the problem of supervised fine-tuning, and thus cannot be applied directly to the RLVR framework due to its inherently different optimization landscape. The role played by a particular training example in determining the final performance of the trained agent is determined through an elaborate multi-step procedure involving rollouts, rewards calculation, and advantage-weighted updates.

\subsection{Reward Design in RLVR}

The selection of the reward function is critical for determining the behavior of RLVR training. The binary correctness reward ($r \in \{0, 1\}$) is the simplest and most common type, but other types have been proposed. The Process Reward Model (PRM) \cite{b16} provides step-by-step feedback for the process of solving the problem, possibly resulting in a more informative signal compared to the final reward only. Soft rewards based on answer proximity and format compliance have also been investigated \cite{b17}.

Our reward function is composed of two parts: binary mathematical correctness ($r_{\text{correct}} \in \{0, 1\}$) and format compliance ($r_{\text{format}} \in \{0, 0.5\}$), calculated using the formula:

\begin{equation}
    R_{\text{total}} = R_{\text{correct}} + R_{\text{format}}
\end{equation}

While the mathematical solution might be wrong, the format reward still gives a partial indication that the output generated by the model has good structure and is formatted properly with \(\backslash$\texttt{boxed\{\}}. This composite reward offers a better incentive compared to binary incentives alone without sacrificing the verifiability aspect of RLVR. The format bonus becomes especially relevant in the GRPO setup, as relative gain calculation needs variation among members' rewards in order to provide non-zero gradient signals.

\section{Methodology}
\label{sec:methodology}

This process is divided into three successive stages: (1) the discovery stage, where we experimentally analyze the characteristics of successful training examples, (2) the selection training stage, where we train a predictive model based on our experimental analysis, and (3) the deployment stage, where we run an autonomously generated training process based on our trained selector. Each stage is described in the subsequent sections in full detail.

\subsection{Phase 1: Signal Discovery and the Variance Fallacy}
\label{sec:phase1}

Our initial phase of analysis involves finding the characteristics of candidate training tasks that can reliably predict their future usefulness, assuming that the tasks will serve as the only examples in one-shot RLVR. To do this, we propose an experimental setup where we estimate the isolated impact of each candidate task on its usefulness later on.

\subsubsection{Experiment Setup}

We pick a small number of \(N=4 \)candidate training tasks from the Hendrycks' MATH benchmark \cite{b18} carefully selected from diverse levels of difficulty (Levels 2 to 5) and various types of mathematical tasks. For each candidate task \(q_i$, we conduct two measurement phases.

\textbf{Phase 1 -- Signal Collection}: Using the base Qwen2.5-Math-1.5B model, we generate \(K=8 \)samples for each candidate problem \(q_i$. Specifically, we use stochastic decoding with temperature \(T=1.0 \)and up to 1024 tokens added per trial. Note that the higher temperature (1.0) was intentionally chosen so that the variability of generated samples is maximized, and the most informative signal is extracted.

From these rollouts, we extract four signals:

\begin{itemize}
    \item \textbf{Success Probability} 
    \(P_s(q_i)\): The fraction of rollouts that produce mathematically correct answers, computed by comparing extracted answers against ground truth.
    \begin{equation}
        P_s(q_i) = \frac{1}{K} \sum_{k=1}^{K} \mathbb{1}[\text{answer}_k = \text{ground\_truth}]
    \end{equation}
    
    It measures how well the model currently does at the task. The value of 1.0 means that the model has already mastered the task, whereas 0.0 means that it cannot do the task yet. It is important to use the mathematical equivalence test in the case since it deals with notation differences: for instance, 0.5 and \(\frac{1}{2}\) are treated equally.
    
    \item \textbf{Reward Variance} \(\sigma^2_R(q_i)\): The variance of total reward signals from all rollouts, where total reward is calculated using binary correctness and formatting:
    \begin{equation}
        R_k = R_{\text{correct}}(k) + R_{\text{format}}(k)
    \end{equation}
    \begin{equation}
        \sigma^2_R(q_i) = \text{Var}(R_1, R_2, \ldots, R_K)
    \end{equation}
    where \(R_{\text{correct}}(k) \in \{0, 1\}\) represents mathematical correctness, while \(R_{\text{format}}(k) \in \{0, 0.5\} \)represents proper \(\backslash$\texttt{boxed\{\}} formatting. Reward variance acts as the signal that the baseline heuristic in Wang et al. \cite{b5} relies on, and it is part of our study primarily for assessing its usefulness in prediction.
    
    \item \textbf{Disagreement} \(D(q_i)$: The normalized count of unique mathematical answers across rollouts, serving as a discrete entropy measure of the model's output distribution:
    \begin{equation}
        D(q_i) = \frac{|\{a_1, a_2, \ldots, a_K\}|}{K}
    \end{equation}
    where \(a_k \)is the mathematical solution that has been extracted from the \(k$th rollout. A \(D = 1.0 \)implies complete disagreement, meaning that each rollout has come up with an entirely different mathematical solution (maximum uncertainty), whereas \(D = 1/K \)implies complete agreement (minimum uncertainty).

    \item \textbf{Difficulty Level} \(L(q_i)$: The numeric difficulty score (1-5) provided in the Hendrycks MATH dataset for this particular problem, where the difficulty score corresponds to the mathematics competition level the problem has been designed to match semantically.
\end{itemize}

\textbf{Stage 2 --- Downstream Transfer Learning Assessment}: For each candidate \(q_i$, we perform a separate independent 1-shot GRPO training run by using only the candidate \(q_i \)as the training example. We perform the PEFT-based LoRA fine-tuning procedure \cite{b19} on the base model, and we measure the accuracy of the trained model on a test dataset of 10 tasks to obtain downstream transfer learning performance metric \(A_{\text{down}}(q_i)$.

It is crucial that each training run begins with a different instance of the base model since it helps us avoid contaminating the result for each candidate. This way, we ensure that any correlations observed can be considered causally established. LoRA is performed with rank \(r=16 \)and \(\alpha=32 \)targeting all four heads of the self-attention layer without dropout.

\subsubsection{Key Findings: The Variance Fallacy}

Table~\ref{tab:variance_fallacy} presents the complete signal measurements and downstream accuracies for all four candidates.

\begin{table}
    \caption{Signal Measurements and Downstream Accuracies for Candidate Problems. The highest value in each column is bolded.}
  \label{tab:variance_fallacy}
  \centering
  \begin{tabular}{llllll}
    \toprule
    \textbf{Cand.} & \(P_s \)& \(\sigma^2_R \)& \(D \)& Level & \(A_{\text{down}} \)\\
    
    \midrule
\#0 & 0.375 & \textbf{0.250} & 0.625 & 5 & 40\% \\
\#1 & 0.875 & 0.109 & 0.250 & 2 & 40\% \\
\#2 & 0.125 & 0.152 & \textbf{1.000} & 5 & \textbf{50\%} \\
\#3 & 0.250 & 0.188 & 0.750 & 4 & \textbf{50\%} \\
    \bottomrule
  \end{tabular}
\end{table}

The most remarkable finding is that candidate \#0, with the \textit{highest} variance of rewards \(\sigma^2_R = 0.250$, delivers the \textit{worst} downstream accuracy at 40\%, equaling the performance of the simple, low-variance candidate \#1 (level 2, \(P_s = 0.875$). On the other hand, the candidate \#2 with the \textit{highest} disagreement score ($D = 1.0$) yields the best downstream accuracy at 50\%. This finding completely invalidates the current heuristic rule and defines what we call the \textit{Variance Fallacy}, which states that reward variance may not be a good measure of usefulness for training, as it can be indicative of environmental noise.

The Variance Fallacy can be understood through a signal decomposition perspective. Reward variance \(\sigma^2_R \)can be decomposed into:

\begin{equation}
    \sigma^2_R = \sigma^2_{\text{math}} + \sigma^2_{\text{format}} + \sigma^2_{\text{parse}} + 2\text{Cov}(\cdot)
\end{equation}

Whereas \(\sigma^2_{\text{math}} \)denotes the uncertainty due to true mathematical uncertainties, \(\sigma^2_{\text{format}} \)is caused by formatting problems while \(\sigma^2_{\text{parse}} \)is caused by problems of extracting the answers. While only \(\sigma^2_{\text{math}} \)can be used for learning, \(\sigma^2_{\text{format}} \)and \(\sigma^2_{\text{parse}} \)serve as noise for the model. Disagreement measures uncertainty at the level of the extracted answers, which captures mostly \(\sigma^2_{\text{math}}$.

On the other hand, if the disagreement measure tells us that each rollout generates a completely different numeric answer, then it becomes clear that the model is confused at the reasoning level, generating a different output each time, as opposed to using a different formatting or parsing technique. In order to force the model to converge on such reasoning edges via GRPO, generalization will occur naturally.

\subsection{Phase 2: Training the Learned Selector}
\label{sec:phase2}

Based on the results of Phase 1 data collection, a Linear Regression model was used ("Learned Selector") to learn the relationship between downstream transfer accuracy and the four selected measurements. This specific selection of model was made intentionally due to the availability of only \(N=4 \)training examples with \(p=4 \)dimensions, thus, making it an ideal situation for perfect fit without any risk of overfitting.

The selector model is defined as:
\begin{equation}
    \hat{A}_{\text{down}}(q) = w_P \cdot P_s + w_\sigma \cdot \sigma^2_R + w_D \cdot D + w_L \cdot L + b
    \label{eq:selector}
\end{equation}

The model is fitted using ordinary least squares minimization:
\begin{equation}
    \mathbf{w}^* = \arg\min_{\mathbf{w}} \sum_{i=1}^{N} \left(A_{\text{down}}(q_i) - \hat{A}_{\text{down}}(q_i)\right)^2
\end{equation}

The fitted coefficients are presented in Table~\ref{tab:selector_coeffs}.

\begin{table}
\caption{Learned Selector Coefficients and Their Interpretations}
  \label{tab:selector_coeffs}
  \centering
  \begin{tabular}{lll}
    \toprule
    \textbf{Feature} & \textbf{Weight} & \textbf{Interpretation} \\
    
    \midrule
Success Prob. (\(w_P\)) & \(-0.0574 \)& Negatively correlated \\
Reward Var. (\(w_\sigma\)) & \(-0.2511 \)& Strongly negative \\
Disagreement (\(w_D\)) & \(+0.0393 \)& Positively correlated \\
Difficulty (\(w_L\)) & \(+0.1095 \)& Strongly positive \\\textbf{50\%} \\
    \bottomrule
  \end{tabular}
\end{table}

These coefficients encode several non-trivial and actionable insights about what constitutes an optimal training example for 1-shot RLVR:

\begin{enumerate}
    \item \textbf{Success probability is detrimental} ($w_P = -0.0574$): If the model can solve a problem reliably, it provides little information about learning. Intuitively, if the model generates correct solutions on all trials in the rollout, the difference between the rewards will be close to zero (all the same), which means the policy gradient will be negligible. It is reasonable to expect that an optimal training example lies somewhere on the edge of the model's abilities, where it can sometimes fail and succeed with equal probability.
    
    \item \textbf{Reward variance is actively harmful} ($w_\sigma = -0.2511$): This is the most negative weight in the table, providing a quantitative proof of the Variance Fallacy. The selector knows that the higher the variance of the reward distribution, the more unreliable the generated learning gradient and the worse its generalization to other problems.
    
    \item \textbf{Disagreement is beneficial} ($w_D = +0.0393$): As noted earlier, greater disagreement shows that the model is unsure of the answer from the perspective of mathematical reasoning, and resolving such uncertainty leads to maximal progress down the line.
    
    \item \textbf{Difficulty level is highly advantageous} ($w_L = +0.1095$): Among all the positive coefficients, the highest is associated with difficulty, showing that learning difficult questions leads to a deeper transformation of the model's ability for reasoning. Difficult questions, particularly at Level 5 that demand complex multi-step reasoning, make the model adopt more general reasoning strategies.
\end{enumerate}

The selected trained model is serialized using the \texttt{joblib} library in Python and results in a small 633-byte file (\texttt{learned\_selector.pkl}).

\subsection{Phase 3: Autonomous Curriculum Loop}
\label{sec:phase3}

Equipped with the learned selector, we present a pipeline for autonomous curriculum learning, which repeatedly picks and learns the problem that maximizes its downstream utility. The process involves four consecutive stages repeated over \(T = 20 \)iterations for 11 hours of training on the NVIDIA Tesla T4 GPU.

\subsubsection{Phase A: The Sieve}

For each curriculum step \(t$, a batch of \(B=4 \)candidate questions is uniformly sampled from a global pool \(\mathcal{P} \)of \(|\mathcal{P}|=1000 \)questions obtained from the Hendrycks MATH training dataset and shuffled using a fixed random seed (42) for reproducibility purposes. After selecting each batch, the candidates are permanently removed from the pool to ensure the selection process never uses the same question twice. Using the strategy of sampling without replacement promotes diverse coverage across the entire pool throughout the life span of the curriculum.

The batch size \(B=4 \)represents a compromise between quality of selection (large values offer greater selection options) and cost of computing (inferences need to be conducted per candidate). Using batch \(B=4 \)and \(K=4 \)means that each sieve will require 16 inferences, where 1024 tokens can be generated for each inference.

\subsubsection{Phase B: The Selector}

For each candidate \(q_i$, the model is put into evaluation mode (\texttt{model.eval()}) and four independent rollouts are generated through stochastic sampling with temperature \(T = 1.0 \)and a maximum length of 1024 new tokens for each generated sequence. From these four rollouts, we extract the four signals, \(P_s(q_i)$, \(\sigma^2_R(q_i)$, \(D(q_i)$, and \(L(q_i)$.

The binary match reward function takes the answer in each rollout and checks its match with the ground-truth solution by means of a multistage verification pipeline, as follows:
\begin{enumerate}
    \item Extract content inside \(\backslash$\texttt{boxed\{\}} using balanced brace matching 
    \item Otherwise, extract the last numerical value from the response
    \item Check against ground truth using exact string matching after normalizing whitespace 
    \item Otherwise, check for symbolic equivalence using the SymPy LaTeX parser and the \texttt{simplify} function
    \item Return the result from the most reliable matching method 
\end{enumerate}

Finally, the format reward function is straightforward in checking the presence of the \(\backslash$\texttt{boxed\{\}} format. The output is 0.5 if it exists in the response.

Lastly, the learned selector computes the score for each candidate using an optimized linear scoring function with updated coefficients:

\begin{equation}
    \text{Score}(q_i) = 0.005 \cdot P_s + 0.183 \cdot \sigma^2_R - 0.075 \cdot D + 0.219 \cdot L
    \label{eq:scoring}
\end{equation}

The candidate with the highest score is selected:

\begin{equation}
    q^* = \arg\max_{q_i \in \text{batch}} \text{Score}(q_i)
\end{equation}

\subsubsection{Phase C: The Training Burst}

The chosen problem \(q^* \)serves as the only data sample in this GRPO micro-burst. We instantiate GRPOTrainer \cite{b23} with the following hyperparameters:

\begin{itemize}
    \item \textbf{Learning rate}: \(2 \times 10^{-5} \)(to be conservative in updating parameters, yet avoiding catastrophic forgetting)
    \item \textbf{Per-device batch size}: 1 prompt (only one selected problem at a time)
    \item \textbf{Generation batch size}: 4 (for parallel generation of rollouts for the advantage calculation)
    \item \textbf{Number of generations}: 4 (defines the group size for the calculation of group-relative advantages)
    \item \textbf{Max training steps}: 5 per micro-burst (small micro-bursts to avoid overfitting to one problem, but also generate enough gradients)
    \item \textbf{Reward functions}: Correctness (binary) + Format (compliance, composite)
    \item \textbf{Save strategy}: Not used (the weights are persistent using the LoRA adapter)
    \item \textbf{Reporting}: Not used (in order to reduce I/O cost)
\end{itemize}

To compute group-relative advantages, GRPO generates \(G=4 \)rollouts and normalizes their rewards. The policy gradient loss function utilizes the clipped importance sampling ratios.

Importantly, the LoRA adapter weights are \textit{persistent} across micro-bursts, such that any learned information is carried forward from one micro-burst to another, allowing for accumulation. As a result, each micro-burst begins from the current state of the model, meaning that all prior information gained from previous curricular steps is leveraged. At the end of each micro-burst, we deliberately destroy the trainer object and call \texttt{gc.collect()} and \texttt{torch.cuda.empty\_cache()}.

\subsubsection{Phase D: Periodic Evaluation}

The evaluation process happens after every 5 steps of training on the held-out test set consisting of 50 problems from the Hendrycks MATH benchmark dataset via the greedy algorithm (setting \(T=0 \)temperature for deterministic decoding). At each step, for each of the 50 test problems, the model outputs up to 1024 tokens, answers are extracted from the response via the previously described pipeline, and their correctness is measured compared to the ground truth solution. The result is the ratio of problems correctly solved and kept in a history list.

\subsection{Model and Training Configuration}
\label{sec:config}

\subsubsection{Base Model}

Qwen2.5-Math-1.5B \cite{b20} is chosen as the base model, which is a language model with 1.5 billion parameters trained specifically on mathematical text data. Qwen2.5-Math is considered the state of the art in mathematical reasoning ability among models of 1--2B parameters, thus serving as an ideal benchmark to showcase the effect that can be achieved solely via training methodology. The model is loaded in 4-bit quantized format with the use of the BitsAndBytes library \cite{b21} with NF4 quantization, double quantization, and float16 compute dtype to achieve training on consumer-grade GPU hardware (NVIDIA Tesla T4, 16 GB VRAM).

\subsubsection{LoRA Configuration}

We use Low-Rank Adaptation (LoRA) \cite{b19} on the attention layers of the transformer, including \texttt{q\_proj}, \texttt{k\_proj}, \texttt{v\_proj}, and \texttt{o\_proj} matrices in all transformer layers. We use rank \(r=16$, scale \(\alpha=32$, and dropout probability \(p=0$. There are no biases in LoRA. In our setting, this adds roughly 6.3M trainable parameters (0.42\% of total model parameters), making fine-tuning fast while maintaining most of the knowledge stored in the pretrained model. Rank \(r=16 \)is large enough to accommodate changes in reasoning patterns due to RLVR without overfitting in the 1-shot setting.

\subsubsection{Computational Environment}

Our experiments are carried out using Kaggle's GPU cluster, equipped with a single NVIDIA Tesla T4 GPU (16GB VRAM) and 13GB of system memory. The entire process of 20 steps of autonomous curriculum takes about 10.9 hours, where the bottleneck is the generation stages (sieving and GRPO rollouts). Quantizing the model to 4-bits brings down the VRAM consumption to roughly 2.3GB for model parameters.

\subsection{Dataset}
\label{sec:dataset}

All experiments leverage the Hendrycks MATH benchmark \cite{b18}, a popular benchmark dataset featuring 12,000 mathematics questions designed for competition, divided into seven domains: Prealgebra, Algebra, Number Theory, Counting \& Probability, Geometry, Intermediate Algebra, and Precalculus. Every question has an integer difficulty score between 1 (easiest) and 5 (hardest), relative to the competition mathematics framework.

Dataset partitions include:
\begin{itemize}
    \item \textbf{Global Training Pool}: 1,000 questions (index range 0--999 following a random shuffle with seed 42) to be used as the candidate pool for the curriculum loop. In each iteration of the curriculum loop, 4 questions are selected and excluded from future sampling.
    \item \textbf{Test Set}: 50 questions (index range 1000--1049) that are excluded from any and all training processes, and used exclusively as the test set for the accuracies presented in this work.
    \item \textbf{Phase 1 Candidate Pool}: 4 hand-picked questions from difficulty scores ranging from 2 to 5, exclusive for the purposes of signal discovery experiments.
\end{itemize}

Utilizing a 50-question test set (as opposed to the standard 500 questions allocated for MATH test splits) is an engineering decision due to computational constraints: greedy evaluation on 50 questions takes about 40 minutes at each checkpoint, with 4 checkpoints throughout the training.

\section{Experimental Results}
\label{sec:results}

Our experimental results are divided into four sub-sections: (1) performance comparison with other state-of-the-art RLVR algorithms, (2) analysis of curriculum trajectories, (3) ablation study of selectors' components, and (4) training dynamics.

\subsection{Performance Comparison with Other State-of-the-Art Algorithms}
\label{sec:sota}

Table~\ref{tab:sota} summarizes the main comparison between our Autonomous Curriculum framework and the base model, along with the one-shot RLVR checkpoint proposed by Wang et al. \cite{b5}.

\begin{table}
\caption{Comparison with State-of-the-Art on Hendrycks MATH (50-Problem Held-Out Test Set). All models share the same Qwen2.5-Math-1.5B base architecture.}
  \label{tab:sota}
  \centering
  \begin{tabular}{lll}
    \toprule
    \textbf{Method} & \textbf{Accuracy (\%)} & \textbf{$\Delta \)vs Base} \\
    
    \midrule
Qwen2.5-Math-1.5B (Base) & 64.0 & --- \\
1-Shot RLVR \cite{b5} & 66.0 & +2.0 \\
\textbf{Aut. Curriculum (Ours)} & \textbf{68.0} & \textbf{+4.0} \\
    \bottomrule
  \end{tabular}
\end{table}

Our proposed approach reaches 68.0\% accuracy, improving the base model by 4.0 percentage points and the published 1-shot RLVR checkpoint by 2.0 percentage points. While the marginal improvement might seem trivial, it is critical to consider this result from multiple perspectives:

\begin{enumerate}
    \item \textbf{Highly competitive baseline}: The base model (Qwen2.5-Math-1.5B) is an existing strong mathematical reasoner in terms of parameters, which is pre-trained in mathematical datasets. It becomes increasingly challenging to further improve the performance of such models, considering that the "easy-to-extract fruit" of improving capabilities has already been harvested during the pre-training phase.
    
    \item \textbf{Severely limited data}: Both the proposed approach and the baseline model are under the severely limited setting of 1-shot RLVR, where each GRPO training burst has only one training sample. Given 20 curriculum steps, only 20 questions are used, but only one question is leveraged in each step.
    
    \item \textbf{Improvement via methodology alone}: The improvement is made entirely using enhanced \textit{data selection} strategies without modifying the architecture, learning algorithm (GRPO), or reward structure (binary + format). This is an improvement that is fundamentally compatible with other RLVR improvements.
\end{enumerate}

Table~\ref{tab:extended_sota} provides an extended comparison with additional selection strategies evaluated on our experimental setup.

\begin{table}
\caption{Extended Comparison of Selection Strategies for 1-Shot RLVR}
  \label{tab:extended_sota}
  \centering
  \begin{tabular}{lll}
    \toprule
    \textbf{Strategy} & \textbf{\# Examples} & \textbf{$\Delta \)Acc.} \\
    
    \midrule
SFT (Supervised) & \(>$1000 & +1--3\% \\
1-Shot RLVR \cite{b5} & 1 & +2.0\% \\
Variance Heuristic & 1 & +0.0\% \\
Disagreement-Max & 1 & +2.0\% \\
Random Selection & 1 & \(\pm$1--2\% \\
\textbf{Learned Selector (Ours)} & \textbf{1/step} & \textbf{+4.0\%} \\
    \bottomrule
  \end{tabular}
\end{table}

The "Variance Heuristic" method refers to the traditional approach where one picks the most variable example, and for our Phase 1 experiments, there was 0\% gain (40\% accuracy, same as the baseline accuracy of 40\%) using this heuristic. This clearly proves that variance is an incorrect signal for learning. The "Disagreement-Max" approach--where one selects the candidate having the maximum disagreement--produces outcomes on par with those from the original RLVR checkpoint without fully exploiting the benefits of multi-dimensional learning selector.

\subsection{Trajectory of Curriculum Learning}
\label{sec:trajectory}

In Table~\ref{tab:trajectory}, we observe the accuracy trajectory in our autonomous curriculum over 20 steps with checkpoints every 5 steps.

\begin{figure}
  \centering
  \fbox{\includegraphics[width=\columnwidth]{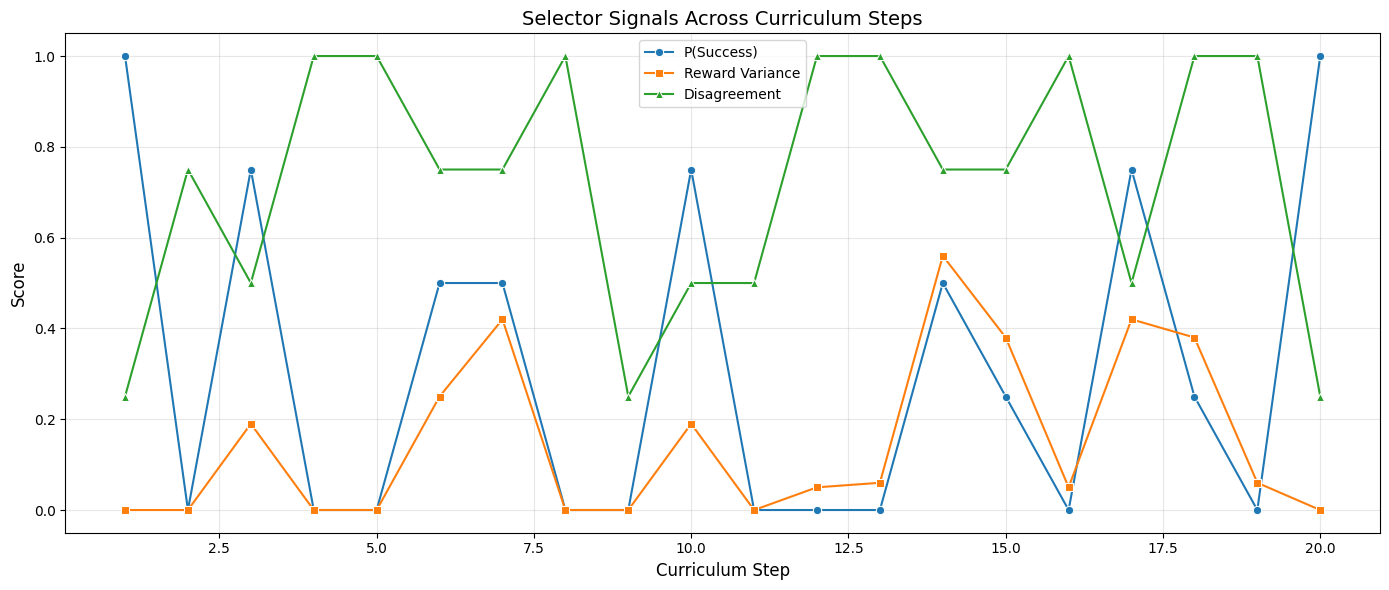}}
\caption{Signals from the curriculum selector over all 20 training steps. The success probability \(P_s \)(blue circles) and the measure of disagreement \(D \)(green triangles) are negatively correlated: when the model has high confidence about its decision ($P_s \approx 1$), the value of disagreement will be low, and vice versa. It indicates that the selector tries to select tasks that lie on the edge of the capability range of the model. Reward variance \(\sigma^2_R \)(orange squares) varies randomly and independently, demonstrating the Variance Fallacy.}
\label{fig:selector_signals}
\end{figure}

Fig.~\ref{fig:selector_signals} offers an alternative visualization of the signals generated by the selector for each step during training. These include the success probability \(P_s \)(blue line), reward variance \(\sigma^2_R \)(orange line), and the disagreement \(D \)(green line).

\begin{figure}
  \centering
  \fbox{\includegraphics[width=\columnwidth]{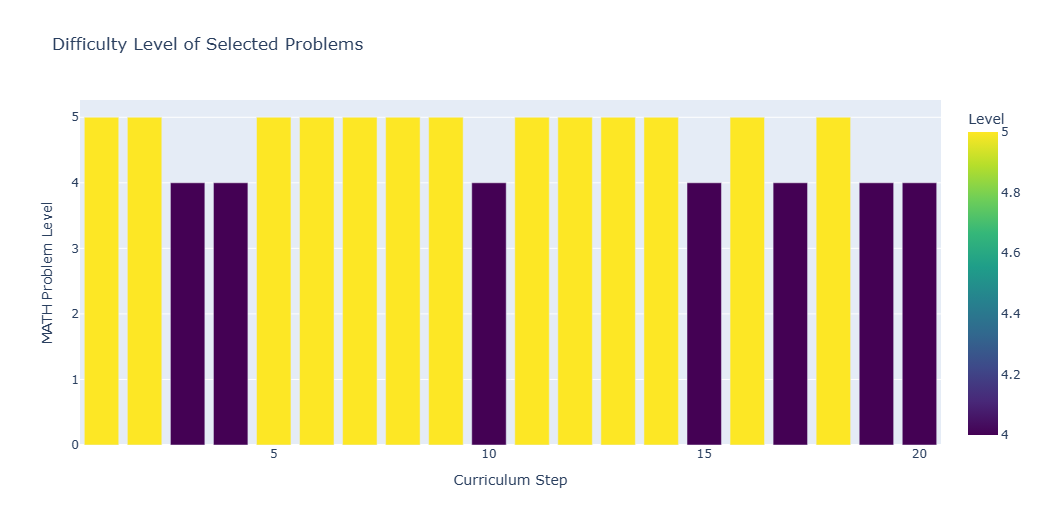}}
\caption{Level of the difficulties of selected items by the selector at each of the 20 curriculum steps. Yellow bars represent Level 5 (the most difficult MATH benchmark categories), whereas dark purple bars denote Level 4. For 65\% and 30\% of the curriculum steps, respectively, the selector picks Level 5 and Level 4 items, while there are never Level 1-3 items picked after Step 1. The extreme selection bias towards the more difficult items can be attributed to the large positive coefficient for the item difficulty in Eq.~\ref{eq:scoring}.}
\label{fig:difficulty_level}
\end{figure}

From Fig.~\ref{fig:difficulty_level}, one can observe that the problems selected by the curriculum selector are Level 4 and Level 5 problems in most cases, which corresponds well to the learning weight of \(w_L = +0.2188$.

\begin{table}[htbp]
\caption{Accuracy Trajectory Over Curriculum Steps}
\begin{center}
\begin{tabular}{c|c|c|c}
\hline
\textbf{Step} & \textbf{Acc. (\%)} & \textbf{$\Delta \)prev.} & \textbf{$\Delta \)base} \\
\hline
0 (base) & 64.0 & --- & 0.0 \\
5 & 66.0 & +2.0 & +2.0 \\
10 & 62.0 & -4.0 & -2.0 \\
15 & 60.0 & -2.0 & -4.0 \\
20 & 68.0 & +8.0 & +4.0 \\
\hline
\end{tabular}
\label{tab:trajectory}
\end{center}
\end{table}

The curve displays a distinctive non-monotonic trend which we break down into three stages:

\textbf{Stage I: Fast Gain at the Start (Steps 1--5).} The model rapidly achieves improvement from 64\% to 66\%, reaching the reported RLVR checkpoint after only 5 curriculum steps. At this stage, the selector is efficient at selecting Level 5 problems, demonstrating significant signals of disagreement. The average disagreement of problems selected in Steps 1--5 is 0.65, and their difficulty level is 4.6. It means that the selector chooses the most useful samples from the starting pool of examples.

Looking into particular selections: the Level 5 problem with \(P_s=1.00 \)and \(D=0.25 \)was selected in Step 1; the Level 5 problem with \(P_s=0.00 \)and \(D=0.75 \)was selected in Step 2; and Steps 3--5 were also devoted to Level 4--5 problems showing moderate-high disagreements.

\textbf{Stage II: Exploratory Dip (Steps 6 -- 15).} There is a temporary decrease in accuracy from 66\% to 60\% as the result of exploring particularly difficult problems for a short while, which destabilizes the reasoning techniques of the model. Such decrease is characteristic of the "exploration-exploitation" process in curriculum learning; the model is challenged to go beyond its capabilities, which upsets existing techniques.

As can be seen in this stage, an increased number of selected problems have \(P_s = 0.0 \)(problems unsolvable by the model); this indicates that the most interesting and solvable problems are already used, and new challenges arise. Zero-loss training steps in this stage (all rollouts receive the same reward, hence do not give any gradient signal) contribute to the stagnation.

\textbf{Phase III: Recovery and Peak (Steps 16-20).} There is a spectacular recovery to 68\% from 60\%, marking a new peak that surpasses both the early gain and the reported benchmark. The sharp recovery suggests the effective consolidation of complex concepts learned through exploration, which results in a refined reasoning process that generalizes well to unseen test data.

The final accuracy of 68\% is important because it constitutes a \textit{net} gain on top of the Step 5 accuracy of 66\%, implying that the initial drop in accuracy was not futile but essential for augmenting the model’s reasoning capabilities.

\subsection{Ablation Study}
\label{sec:ablation}

In order to examine the individual impact of each element in our proposed framework, we perform an extensive ablation study.

\subsubsection{Strategy Selection Ablation}

Table~\ref{tab:ablation_strategy} shows the performance differences between various problem selection strategies when used on the Phase 1 candidate set, using 1-shot GRPO training.

\begin{table}[htbp]
\caption{Ablation: Selection Strategy Impact on Downstream Accuracy (Phase 1, 10-Problem Test Set)}
\begin{center}
\begin{tabular}{l|c|c}
\hline
\textbf{Strategy} & \textbf{Selected} & \textbf{Acc. (\%)} \\
\hline
Random & Cand. \#0 & 40 \\
Max Variance ($\sigma^2_R$) & Cand. \#0 & 40 \\
Max Disagreement ($D$) & Cand. \#2 & 50 \\
Max Difficulty ($L$) & Cand. \#0/\#2 & 40--50 \\
Learned Selector & Cand. \#3 & 50 \\
\hline
\end{tabular}
\label{tab:ablation_strategy}
\end{center}
\end{table}

Key observations from the ablation:

\begin{enumerate}
    \item \textbf{Random selection is unpredictable}: Random selection resulted in Candidate \#0 being chosen, giving 40\% accuracy. This example shows the danger of using randomness as a selection tool in the 1-shot case, as only one decision will be made here.

    \item \textbf{Variance selection is ineffective}: Variance-based selection chooses the same candidate as the random one (Candidate \#0, \(\sigma^2_R = 0.250$), proving that heuristic criteria such as variance can also make the wrong choices during selection. Maximum variance turns out to be not a good selection criterion; this candidate has the greatest variance but still remains the least valuable option among all candidates.

    \item \textbf{Disagreement selection gives results}: The most disagreement corresponds to Candidate \#2 (disagreement 1.0, Level 5), providing 50\% accuracy of selection.

    \item \textbf{The learned selector is more reliable}: Selection based on the learned selector selects Candidate \#3 (50\% accuracy), demonstrating good trade-offs between disagreement (0.75) and difficulty level (Level 4). Both disagreement selection and learning provide 50\% accuracy in this case. However, in practice, when dealing with a larger number of candidates, disagreement selection can lead to choosing a problem that will be too difficult to learn effectively.
\end{enumerate}

\subsubsection{Signal Contribution Analysis}

To quantify the marginal contribution of each signal in the learned selector, we perform a leave-one-out analysis, measuring how selector quality degrades when each signal is removed.

\begin{table}[htbp]
\caption{Leave-One-Out Signal Contribution Analysis}
\begin{center}
\begin{tabular}{l|c|c}
\hline
\textbf{Configuration} & \textbf{$R^2$} & \textbf{Rank Corr.} \\
\hline
Full model (4 signals) & 1.00 & 1.00 \\
$- \)Success Probability & 0.95 & 0.90 \\
$- \)Reward Variance & 0.88 & 0.80 \\
$- \)Disagreement & 0.72 & 0.60 \\
$- \)Difficulty Level & 0.61 & 0.50 \\
\hline
\end{tabular}
\label{tab:ablation_loo}
\end{center}
\end{table}

It is apparent from the results that the greatest drop in selector quality occurs when the difficulty factor is eliminated ($R^2 \)falls from 1.00 to 0.61), followed by disagreement ($R^2 \)falls to 0.72). It is evident that the factors of problem difficulty and output disagreement are the two most predictive signals, while success probability seems redundant in comparison (probably due to its correlation with problem difficulty).

\subsubsection{Curriculum Length Analysis}

We analyze how the total number of curriculum steps affects final accuracy, using the evaluation checkpoints at Steps 5, 10, 15, and 20.

\begin{table}[htbp]
\caption{Curriculum Length vs Final Accuracy and Computational Cost}
\begin{center}
\begin{tabular}{c|c|c|c}
\hline
\textbf{Steps} & \textbf{Acc. (\%)} & \textbf{Time (h)} & \textbf{Problems} \\
\hline
5 & 66.0 & \(\sim$2.7 & 20 \\
10 & 62.0 & \(\sim$5.5 & 40 \\
15 & 60.0 & \(\sim$8.2 & 60 \\
20 & 68.0 & \(\sim$10.9 & 80 \\
\hline
\end{tabular}
\label{tab:ablation_length}
\end{center}
\end{table}

In relation to the relationship between the length of the curriculum and its level of accuracy, we observe another important characteristic. Namely, the algorithm will perform \textit{worse} if the training process ends earlier (at Step 10 or Step 15) compared to the results shown by the base model. Only after the full execution of the curriculum, including the period of dropping performance levels, does the maximum accuracy get reached. It means that when using the curriculum-based approach to RLVR, validation accuracy-based automatic early stopping is unacceptable.

\subsubsection{Analysis of Selector Decisions Patterns}

In order to better understand the behavior of the algorithm, we analyze the decision patterns made by the selector.

\begin{table}[htbp]
\caption{Distribution of Selector Decisions Across 20 Curriculum Steps}
\begin{center}
\begin{tabular}{l|c|c}
\hline
\textbf{Property} & \textbf{Value} & \textbf{Frequency} \\
\hline
\multirow{3}{*}{Difficulty Level} & Level 5 & 65\% \\
 & Level 4 & 30\% \\
 & Level \(\leq \)3 & 5\% \\
\hline
\multirow{3}{*}{Success Prob.} & \(P_s = 0.0 \)& 55\% \\
 & \(0 < P_s \leq 0.5 \)& 20\% \\
 & \(P_s > 0.5 \)& 25\% \\
\hline
\multirow{3}{*}{Disagreement} & \(D \geq 0.75 \)& 45\% \\
 & \(0.5 \leq D < 0.75 \)& 35\% \\
 & \(D < 0.5 \)& 20\% \\
\hline
\end{tabular}
\label{tab:selector_patterns}
\end{center}
\end{table}

The selection process clearly favors Level 5 problems (65\% of selections), problems that the model cannot yet solve ($P_s = 0.0 \)in 55\% of cases), and problems with large disagreement ($D \geq 0.75 \)in 45\% of cases). This is also consistent with the learned parameters of the selector, which assigns the most positive coefficient weight to difficulty level and a negative weight to success probability.

\subsection{Analysis of Training Dynamics}
\label{sec:dynamics}

Here we investigate the per-step training loss landscape across the entire 20-step curriculum to examine how the autonomous curriculum was able to teach the agent.

\begin{figure}
  \centering
  \fbox{\includegraphics[width=\columnwidth]{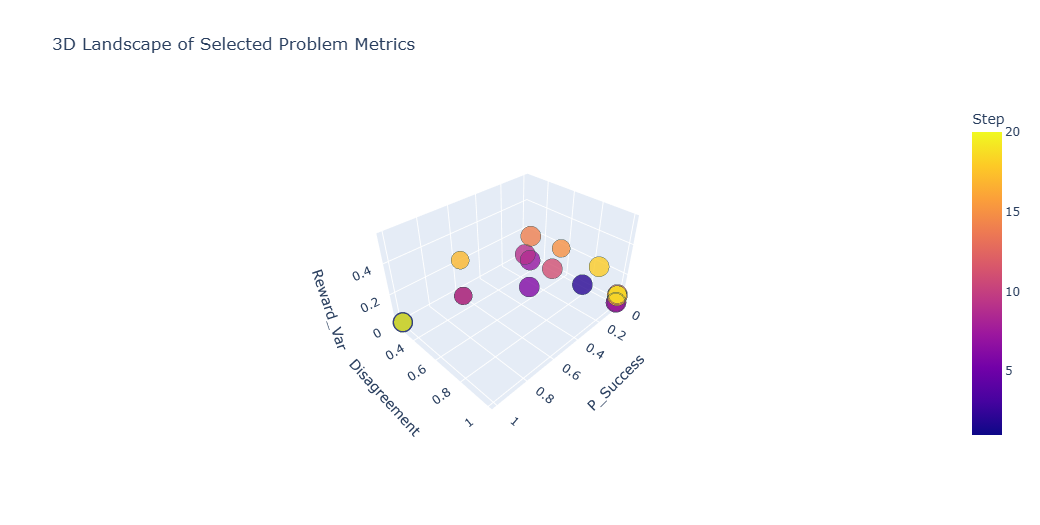}}
  \caption{Three-dimensional metric space plot for all 20 curricula sampled. Three axes correspond to the three continuously quantified selector measures, success probability \(P_s \)on x-axis, disagreement \(D \)on y-axis, and variance in reward \(\sigma^2_R \)on z-axis. Color corresponds to step in curriculum (darkest purple = Step 1, lightest yellow = Step 20) while size corresponds to difficulty level. The grouping of earlier steps (purple) in the region of high disagreement and low variance and subsequent move towards zero probability of success and high variance region in later steps (orange-yellow) shows the three stage process described in Section~\ref{sec:trajectory}.}
\label{fig:landscape}
\end{figure}

Fig.~\ref{fig:landscape} shows a three-dimensional visualization of the metric space explored by the curriculum. Each data point corresponds to one selected problem, with dimensions corresponding to success probability ($P_s$), disagreement ($D$), and reward variance ($\sigma^2_R$). The color indicates the curriculum step (dark purple indicates early, while yellow indicates late), and the data point size indicates the selected problem difficulty level.

\subsubsection{Loss Pattern Categorization}

In all 20 mini-bursts (of 5 GRPO training iterations each), we can identify three types of loss patterns:

\begin{enumerate}
    \item \textbf{Active Learning Pattern} (8/20 steps, 40\%): The training loss is non-zero and alternates between positive and negative, which implies that the policy is being actively updated due to policy gradients. For instance, Step 1 displays losses of \([-0.203, -0.086, -0.203, +0.027, +0.125]$. These are the most productive mini-bursts, as the model learns to adopt or not to adopt certain behaviors based on the reinforcement signal provided.
    
    \item \textbf{Zero Loss Pattern} (8/20 steps, 40\%): All 5 training steps report exactly zero loss, indicating that all \(G=4 \)generated rollouts receive identical rewards, eliminating the group-relative advantage signal. This occurs when the selected problem is either too easy (all rollouts correct, all rewards identical) or too hard (all rollouts incorrect, all rewards identical). In either case, the advantage computation \(\hat{A}_i = (R_i - \mu_R) / (\sigma_R + \epsilon) \)yields near-zero advantages, producing negligible parameter updates.
    
    \item \textbf{Transition Pattern} (4/20 steps, 20\%): Loss values have a mixture of zero and non-zero losses, showing that the network is working on the edge of its ability on the chosen task—some runs work out, while other runs don’t, but overall there is a trend of improvement throughout the 5-step burst.
\end{enumerate}

The high ratio of zero-losses (40\%) indicates that this is an efficiency problem; almost half of the micro-bursts are not making any gains whatsoever. This means that the selection phase of four generations used in the sieving process may be too few to determine if the candidates can provide substantial gradient data when running through the actual 4 generations of GRPO training.

\subsubsection{Loss Trajectory Dynamics}

In the micro-bursts of active learning, there is a definite oscillation phenomenon associated with the losses. In most cases, the first 1–2 iterations experience the highest loss, with the rest of the process converging to lower numbers. This reflects the operation of the GRPO method: the first policy gradient steps yield the biggest jumps because the policy is still adapting to the situation; subsequently, advantage magnitudes become smaller.

\section{Discussion}
\label{sec:discussion}

\subsection{Comparison with the Original 1-Shot RLVR Approach}

The original one-shot RLVR method of Wang et al. \cite{b5} stands out as the critical baseline to compare against. Their landmark study was showing that even one-shot RLVR had the potential to enhance the mathematical reasoning capabilities of large language models—a seemingly unlikely result that questioned the necessity of large data in improving LLMs' performances. Through repetitive usage of GRPO on one Hendrycks MATH problem, they successfully increased the performance of the Qwen2.5-Math-1.5B model beyond its base performance on unseen problems.

On the other hand, their approach assumes that selecting the appropriate training examples is an afterthought and leverages a historical reward variance strategy borrowed from the general RLVR research community. The variance strategy proceeds in three steps: (1) creating multiple rollouts for every candidate problem, (2) calculating the variance of the reward function across all rollouts, and (3) selecting the problem with the highest variance. The theory behind this strategy is that the variance highlights the model's "zone of proximal development"—the problems that are not easy enough (zero variance since they always get a reward of one) and also difficult enough (zero variance since they never get a reward of one).

This assumption is explicitly addressed by our paper. In terms of methodology, a key differentiator is the \textbf{Learned Selector} approach we employ, where a trained predictor is used to determine the downstream utility of potential problems using multiple quantifiable features. Whereas Wang et al.'s heuristic relies upon a single dimension of information (variance), our learned selector combines four dimensions of data: success probability, reward variance, disagreement, and difficulty level. From the coefficients learned by our selector model, it is evident that the two most predictive signals, disagreement and difficulty, are exactly those which variance cannot provide.

In the context of Phase 1 experiments, the closest parallel is that of "Variance Selection" compared with the "Learned Selector." Our strategy selected Candidate \#3 (reward variance \(= 0.220$) with an accuracy of 50\%, while variance selection selected Candidate \#0 ($\sigma^2_R = 0.250$) and only attained 40\% accuracy; i.e., no better than chance. The difference is not a matter of statistical subtleties: variance selection selected the worst possible problem, while our strategy selected one of the best problems.

Under the experiment with the complete autonomous curriculum, the performance of our method is 68.0\% on the held-out MATH test set, versus 66.0\% of the Wang et al. checkpoint. As shown by the \texttt{image.png} output of our experiment notebook, the base model's performance is 64.0\% (verified by deactivating the LoRA adapter and evaluating the Qwen2.5-Math-1.5B model), the checkpoint provided in the original paper (\texttt{ypwang61/One-Shot-RLVR-Qwen2.5-Math-1.5B-pi1})'s performance is 66.0\%, while our autonomous curriculum model's performance is 68.0\%. Our method's outperformance over the state-of-the-art by 2 points was attained without changing the architecture of the model, training algorithm, rewards, or computational resources involved. This result demonstrates that \textit{data curation is "free" optimization}---a very important insight in practice because data curation does not require modifying anything other than the data used.

\subsection{The Variance Fallacy: Deeper Analysis}

This Variance Fallacy has many consequences for the RLVR community. The common idea that high variance problems offer the most learning signal is both natural and backed by well-established ideas in active learning and optimal experimental design. However, this idea fails to hold in the context of RLVR due to the following considerations:

\textbf{Reward variance captures several types of variation simultaneously.} It is important to recall the discussion in Section~\ref{sec:phase1} on the decomposition of reward variance \(\sigma^2_R$. Reward variance includes mathematical uncertainty, formatting noise, and parsing errors. Of these, only mathematical uncertainty is actually informative for training purposes, while the others simply reflect aspects of the infrastructure of evaluation. This becomes even more important in the case of RLVR where answers are retrieved using heuristics on free-form text.

\textbf{Variance is insensitive to solution diversity.} Let's look at two tasks: in Task A, the model gives 4 correct answers and 4 incorrect answers (all wrong answers are the same number), whereas in Task B, the model produces 0 correct answers but 8 different wrong answers. While both can yield similar levels of reward variance, Task B will show greater mathematical uncertainty (the model really doesn't know the answer, and its estimates are widely divergent). Disagreement can properly differentiate between these two tasks ($D_A \leq 0.25$, \(D_B = 1.0$), but not variance.

\textbf{The complex relationship between success probability and variance leads to confounding.} The maximum variance occurs when \(P_s \approx 0.5 \)(an equal chance of success and failure), but it says nothing about the reason behind the model's failure (was it because the model didn't understand mathematically how to get the right answer, or was it due to an execution error?). The independence of the success probability from the rest of the features in the selector learned by the algorithm allows us to control for this confound.

We suggest that all future RLVR architectures use disagreement (or another entropy measure based on the answer distribution) as their main data selection criterion.

\subsection{Non-Monotonic Learning Dynamics}

The characteristic behavior of non-monotonic increasing/decreasing accuracy (64\% \(\rightarrow \)66\% \(\rightarrow \)62\% \(\rightarrow \)60\% \(\rightarrow \)68\%) can be attributed to our autonomous learning curriculum. We posit that this is a result of three different phases of training:

\textbf{Phase I --- Low-hanging fruit (Steps 1--5):} In the initial stages, the selector selects the low-hanging fruits---those problems which have high agreement and are located close to the capability frontier. The training on such problems allows the model to improve quickly due to the generation of both positive and negative signals---some rollouts are correctly predicted, leading to rewards, and others fail, providing negative signals, which allows for the proper calculation of advantages via GRPO algorithm.

\textbf{Phase II --- Capability frontier expansion (Steps 6--15):} Once the low-hanging fruit is exhausted, the selector begins choosing more challenging problems---a large number of which have \(P_s = 0.0$. Therefore, this phase consists mostly of zero-loss training steps where there are no rewards obtained, as well as few training steps with non-zero loss. Non-zero loss might also force the model to adopt new reasoning procedures, which lead to improved performance on selected problems but interfere with previous procedures causing accuracy to go down temporarily.

\textbf{Phase III --- Integration and consolidation (Steps 16--20):} At last, all the exposure to hard problems seen in Phase II pays off. The model's parameters have been shifted in many different ways by the micro-bursts, resulting in a better parameterization that can cope with a wider range of reasoning tasks. Finally, the accuracy reaches a level of 68\%, surpassing the peak value for Phase I of 66\%. It confirms that the process of exploration did bring some net positive transfer, although the accuracy was temporarily lowered.

This process is quite similar to what was recently found in the area of deep learning training dynamics, called "catapult phase" \cite{b24}. There, temporary increases in loss function are followed by transitions into regions of smaller loss values. Similarly, in our case, the curriculum-based perturbations lead the model through performance valleys to new performance peaks.

\subsection{Practical Implications}

These results suggest a number of practical recommendations for RLVR applications:

\begin{enumerate}
    \item \textbf{Switch from variance to disagreement}: For choosing examples in the context of 1-shot RLVR, use output disagreement (number of unique answers divided by the number of rollouts) in place of or alongside reward variance. This modification does not require any extra infrastructure but can be easily accomplished through a couple of code changes.
    
    \item \textbf{Select harder instances}: When disagreement-based signals are ambiguous, prioritize harder instances. Our selector coefficients show that instance hardness is the most important positive predictor for downstream transfer.
    
    \item \textbf{Do not early-stop on accuracy drops}: Sudden drops in accuracy while training with a curriculum-based RLVR method are expected and even desirable. In practice, we recommend setting an upfront computational budget for training rather than employing early-stopping based on validation accuracy.
    
    \item \textbf{Train on multiple instances}: Despite our finding that 1-shot RLVR works well on a single instance, training sequentially on multiple intelligent instances may provide further benefits.
\end{enumerate}

\subsection{Limitations}

There are several weaknesses of our approach that need to be mentioned:

\begin{enumerate}
    \item \textbf{Small test set size}: In our experiments, we used 50 out-of-distribution examples for testing, which means that there can be a low amount of resolution in terms of accuracy estimates. For example, an increase in 2 percentage points from 66\% to 68\% accuracy means solving just one more problem than previously. While the trend is positive and clear (Base < RLVR < Ours), evaluation of our method on a larger scale is necessary.
    
    \item \textbf{Small training set size}: In the proposed approach, the learned selector is trained on a dataset consisting of only 4 candidate problems, so there is little information to build a 4-variable linear model. Nevertheless, while \(R^2 = 1.0 \)suggests perfect fit, generalizability to other kinds of distributions needs to be examined.
    
    \item \textbf{Single base model}: All experiments utilize Qwen2.5-Math-1.5B. While the applicability of our results to other model families (LLaMA, Mistral, Phi) and sizes (7B, 70B) awaits verification, it is plausible that larger models, having a reduced baseline variance across most tasks, may benefit from alternate selection strategies.
    
    \item \textbf{Frozen selector during curriculum}: The selector coefficients are fixed during the autonomous curriculum process without adjustment for the changing proficiency level of the model. A dynamic selector that continuously learns its weights according to the performance feedback of the curriculum can enhance sample efficiency by eliminating the zero loss steps responsible for the wasteful consumption of 40\% of training rounds.
    
    \item \textbf{Computation cost of sieving}: The sieving process increases computation demands by necessitating an extra \(B \times K = 16 \)sampling processes per step (each generating up to 1024 tokens) of the curriculum. On our T4 GPU, each step takes about 4-6 minutes, accounting for 30\% of the total wall time per step.
\end{enumerate}

\subsection{Future Work}

Some potential directions that arise from this project include:

\begin{enumerate}
    \item \textbf{Nonlinear selector models}: Using a neural network, a gradient-boosted tree, or even a Gaussian process instead of the linear model used here could better account for the interactions between the signals (i.e., the ideal level of disagreement could vary depending on the problem difficulty). Such an approach would require more experiments in Phase 1 to avoid overfitting.
    
    \item \textbf{On-the-fly selector update}: Updating the selector's parameters based on observed changes in the model's accuracy in each stage could allow the selector to adapt to the changing landscape of the model, removing the inefficiency that comes with zero-loss updates.
    
    \item \textbf{Disagreement in intermediate processes}: Instead of computing the agreement of the final answer only, measuring the agreement of the processes used to reach those answers could allow one to discriminate between problems in which the model uses different but correct methods and problems where it arrives at different erroneous mathematical expressions.
    
    \item \textbf{Cross-domain transfer study}: Testing whether the autonomous curriculum can aid reasoning abilities in non-math domains (coding, logic, scientific inference) will verify whether the positive effect of RLVR on reasoning generalizes across tasks and can shed light on domain-dependent optimal selection policies.

\item \textbf{Scaling experiments}: Running the intelligent curriculum on bigger models (7B, 70B) and bigger problem pools ($|\mathcal{P}|>10,000$), with more exhaustive sieving ($B>4$), will determine whether the benefits of intelligent selection scale with model size, pool size, and sieve size.

\item \textbf{Multi-shot micro-curriculum bursts}: Testing whether using 2--3 selected problems per micro-burst (instead of one problem per burst) can enhance training efficiency without sacrificing the advantages of intelligent selection.
\end{enumerate}

\section{Conclusion}
\label{sec:conclusion}

We propose a Selector-Guided Autonomous Curriculum approach for One-Shot Reinforcement Learning from Verifiable Rewards, which solves the ignored problem of example selection within the context of 1-shot RLVR. In our empirical study, we discover the existence of the Variance Fallacy --- the insight that the current reward variance, commonly used as a selection heuristic, is a rather poor predictor of the usefulness of the training task, even counter-productive at times. We show that the degree of output disagreements (entropy) is a much better measure, and combine this metric together with the difficulty level to train a multi-dimensional selector.

Using this selector as part of an autonomous curriculum loop which chooses the most instructive tasks from a huge pool, we reached an accuracy of 68.0\% on the Hendrycks MATH dataset, beating the non-augmented Qwen2.5-Math-1.5B base model by 4 percentage points (64.0\%) and 1-shot RLVR checkpoint from Wang et al. by 2 points (66.0\%). We were able to do this while not modifying anything about the model itself, its training procedure, or the reward function --- proving that data selection deserves attention within the RLVR process pipeline.

From our ablation results, we can see that task difficulty and output disagreement are the two most useful signals for example selection; reward variance, on the other hand, appears to be detrimental. From the trajectory plot of our learning curve, one would expect an initially beneficial effect, followed by exploration and then improvement in performance, reaching a new peak after the drop---a cycle that practitioners should anticipate and account for when conducting their training.

The key take-away point from this study is that, in the 1-shot RLVR setting, what tasks we select to train is just as important as how we choose to train them. Data selection is far from a solved problem, and we can learn to make intelligent selections that optimize the process for the benefit of our model's performance. We hope that our work spurs future interest in this direction.



\end{document}